# Zero-Shot Image-Based Large Language Model Approach to Road Pavement Monitoring

Shuoshuo Xu[1], Kai Zhao[1], James Loney[2], Zili Li[1], Andrea Visentin[1]

[1] *School of Civil Engineering, University College Cork, Cork, Ireland*

[2] *Pavement Technology, McCurdy Associates Consulting Engineers Ltd*

**Abstract:** Effective and rapid evaluation of pavement surface condition is critical for prioritizing maintenance, ensuring transportation safety, and minimizing vehicle wear and tear. While conventional manual inspections suffer from subjectivity, existing machine learning-based methods are constrained by their reliance on large and high-quality labelled datasets, which require significant resources and limit adaptability across varied road conditions. The revolutionary advancements in Large Language Models (LLMs) present significant potential for overcoming these challenges. In this study, we propose an innovative automated zero-shot learning approach that leverages the image recognition and natural language understanding capabilities of LLMs to effectively assess road conditions. Multiple LLM-based assessment models were developed, employing prompt engineering strategies aligned with the Pavement Surface Condition Index (PSCI) standards. These models' accuracy and reliability were evaluated against official PSCI results, with an optimized model ultimately selected. Extensive tests benchmarked the optimized model against evaluations from various levels experts using Google Street View road images. The results reveal that the LLM-based approach can effectively assess road conditions, with the optimized model—employing comprehensive and structured prompt engineering strategies—outperforming simpler configurations by achieving high accuracy (1.07 of Mean Absolute Error (MAE) on a 10-point scale) and consistency, even surpassing expert evaluations (MAE of 1.10). Moreover, the successful application of the optimized model to GSV images demonstrates its potential for future city-scale deployments. These findings highlight the transformative potential of LLMs in automating road damage evaluations and underscore the pivotal role of detailed prompt engineering in achieving reliable assessments.

# 1 Introduction

Rapid evaluation of road conditions and timely maintenance can significantly extend the service life of roads, ensure driving comfort and safety, and reduce vehicle wear and repair costs (Abbondati et al. 2021; Aboah and Adu-Gyamfi 2020; Mullaney et al. 2014). Moreover, timely and accurate road assessments are critical for optimizing maintenance prioritization. Traditionally, visual pavement assessments rely on manual inspections using established standards such as the Pavement Condition Index (PCI), Pavement Surface Condition Index (PSCI) (Mullaney et al. 2014), Road Condition Indicator (RCI), and International Roughness Index (IRI). Although these methods are still widely used, they exhibit clear drawbacks—including being time-consuming, subjective, and inconsistent—thereby creating an urgent demand for automated numeric rating systems.

To address the challenges associated with manual evaluations, automated approaches have gained significant attention (Basavaraju et al. 2020; Bhatt et al. 2017; Ouyang et al. 2025), focusing on two main directions: vibration-based methods (Abbondati et al. 2021; Ahmed et al. 2022; Alatoom and Obaidat 2022; Al-Sabaeei et al. 2024; 'Brien et al. 2023; Ouyang et al. 2023) and image-based evaluations (Fu et al. 2024; Ma et al. 2017; Yamaguchi and Mizutani 2024). Vibration-based methods are a cost-effective alternative that utilizes sensors (e.g., accelerometers and gyroscopes) mounted on vehicles to infer road surface irregularities (Dong and Li 2021). While effective in detecting various road anomalies, these methods struggle to differentiate specific distresses, such as cracking or potholes, due to the limited information provided by one-dimensional vibration data (Li et al. 2024; Zhao et al. 2024). Moreover, vibration measurements are highly sensitive to variations in vehicle types and driving behaviors, making standardized analysis difficult. In contrast, image-based evaluations leverage advances in computer vision and deep learning to directly analyze visual features such as cracks, rutting, and potholes (Maeda et al. 2018). Techniques such as object detection, image classification, and semantic segmentation have been extensively explored. For instance, datasets such as Paris-Saclay and Road Quality Dataset have supported the development of convolutional neural networks (CNNs) for crack detection (Jiang et al. 2021; J. Wang et al. 2021), distress quantification, and condition classification (Ma et al. 2017; Qureshi et al. 2023). Advanced models like U-Net and YOLO have further improved accuracy by combining distress identification with severity analysis (Maeda et al. 2018; Majidifard et al. 2020). However, these methods still face challenges, most notably their reliance on large, labelled datasets that are both costly and time-intensive to create. Existing datasets often focus on limited distress types or specific regions, neglecting surface conditions like ravelling and bleeding. In addition, models trained on one dataset frequently struggle to generalize across diverse environmental conditions or different imaging setups, limiting their broader applicability (Arya et al. 2020, 2021).

Large language models (LLMs) have recently garnered considerable attention in civil engineering due to their transformative potential (Chen and Zhang 2024; Liu et al. n.d.; Pu et al. 2024). Initially developed for processing and generating human-like text, LLMs have been successfully adapted for applications such as structural health monitoring, traffic signal control (Movahedi and Choi 2025; Tang et al. 2024), and autonomous driving (Kong et al. 2024). Their strength lies in their ability to generalize across diverse domains without relying on extensive labeled datasets (Bibi et al. 2021). For instance, in pavement condition evaluation, tailored prompt strategies can extract pertinent image features, classify damage severity, and ensure the model's outputs conform to established standards. This adaptability is particularly valuable in civil engineering, where variability in conditions and limited data availability are common challenges (Aragón et al. 2016; Jahan et al. 2023). However, the success of these applications critically depends on prompt engineering—a technique that crafts precise, context-aware instructions to guide the model's outputs.

In this study, we integrate LLMs with image recognition technologies to assess pavement damage severity. By employing various frameworks of prompt engineering, multiple LLM-based models are developed and evaluated against the Pavement Surface Condition Index (PSCI). Based on the optimal model selected, we further explore its potential for application to Google Street View road images. This research underscores the transformative potential of LLMs to automate labour-intensive processes, improve assessment precision, and significantly enhance the efficiency of road condition evaluations, paving the way for broader applications in civil infrastructure management and maintenance.

## 2 LLM-based model building

*2.1 Overall framework*

The overall framework of the proposed method is shown in **Fig.1**, which mainly contains data preparation and model development. Dataset 1 consists of road condition images from the official PSCI documentation (Mullaney et al. 2014) with professionally validated ground truth labels, while Dataset 2 contains images from GSV that were evaluated by experts following PSCI standard. Both datasets were encoded in Base64 format for further processing. The encoded images data was processed through LLMs to obtain road condition ratings, where different LLMs were defined through different prompt engineering techniques. The prompt engineering strategies were designed based on PSCI evaluation standards (section 2.4). Quantitative evaluations were conducted against three benchmarks (road images and condition ratings): official PSCI standards, data extracted from existing literature (Qureshi et al. 2023), and experts' evaluation results. Through comprehensive comparisons, we identified the optimal model configuration for road condition evaluation.

In contrast to conventional approaches requiring data-intensive training (e.g., machine learning (ML) and deep learning (Qureshi et al. 2023)), the proposed method in this study leverages the zero-shot learning capabilities of pre-trained LLMs. By refining prompt engineering strategies to align model outputs with ground truth references, we managed to build an effective LLM-based model without dependence on annotated datasets. This method fundamentally differs from conventional approaches and demonstrates the potential for broader applications in engineering practice. This study focuses on analyzing the applicability of this method in road condition evaluation.

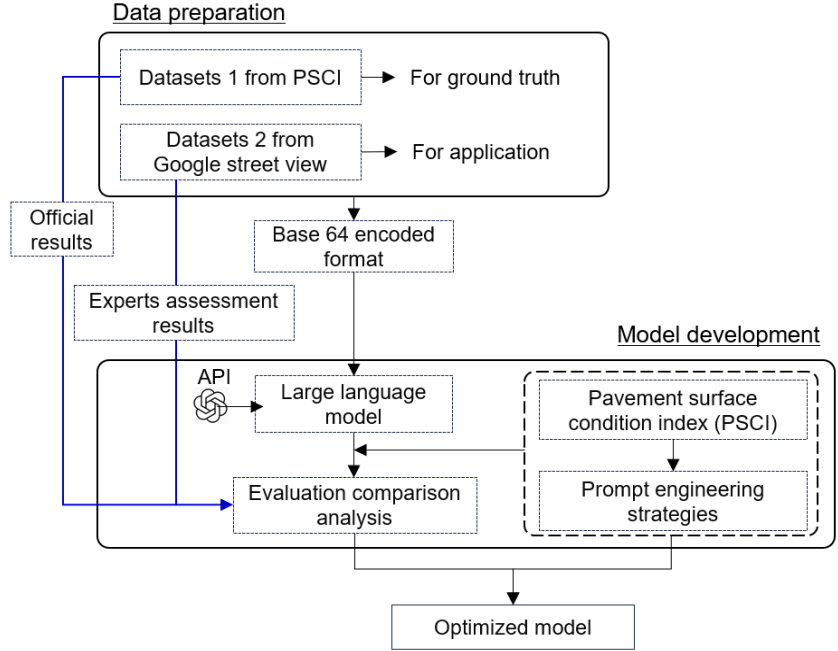

**Figure 1.** Flowchart of building LLMs-based road condition evaluation model

*2.2 Data preparation*

In this study, the selected datasets serve only to validate and evaluate the LLM-based models' results for road condition ratings and are not used for training as in traditional ML methods. The quality of these datasets is critical for model evaluation. We selected data through three sources: (1) official PSCI records (Mullaney et al. 2014), (2) labeled images from existing literature (Qureshi et al. 2023), and (3) diverse Google Street View images, as detailed in **Table 1**.

**Table 1.** Datasets for validation

| Data | Number | Sources | Image size | Road condition ratings |
|---|---|---|---|---|
| Dataset 1 (Mullaney et al. 2014; Qureshi et al. 2023) | 32 | PSCI | 30-800KB | Multiple experts |
| | 10 | Literature | 30-800KB | Multiple experts |
| Dataset 2 | 82 | Selected GSV images | 400-800KB | Three road pavement experts (this study) |

Dataset 1 comprises images certified by PSCI and data from literature. Road ratings for these images are assigned by multiple experts, so the ratings in Dataset 1 serve as the ground truth for our models. Additionally, the number of images selected for each rating level (1–10) is nearly equal (detailed level description in section 2.3), ensuring comprehensive coverage and data balance across all road conditions. As shown in **Fig.2**, the labels represent road ratings and clearly demonstrate a quality improvement with increasing ratings. For example, ratings 9 and 10 correspond to roads that are extremely smooth with almost no surface damage. As the ratings decrease, the roads exhibit progressively more defects, such as cracking, structural damage, potholes, and rutting deformation.

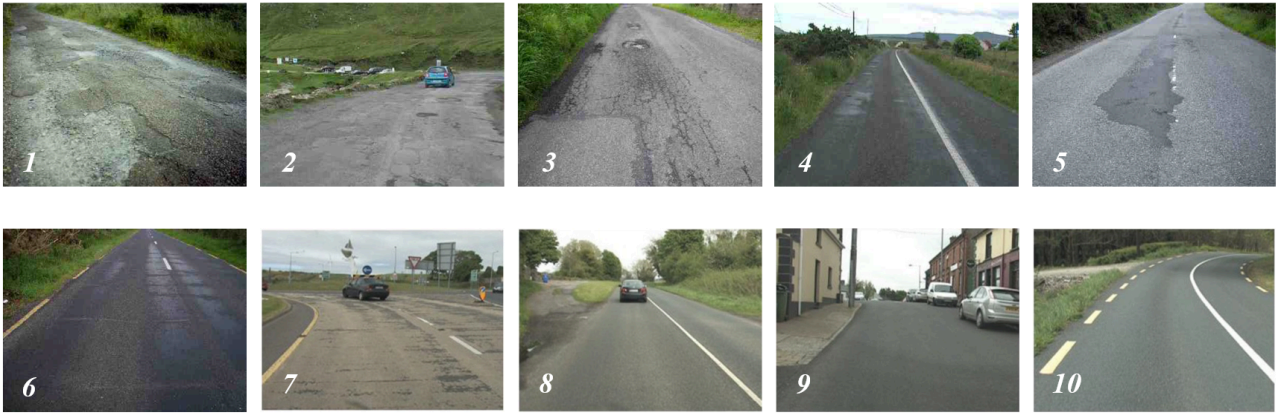

**Figure 2**. Road surface conditions rated from 1 to 10 from PSCI [1,17]

To further validate the applicability of the proposed model, we selected road images from GSV as Dataset 2. This dataset comprises 82 road images, representing broad road conditions, as shown in **Fig.3**. Based on partial GPS data, these images were directly acquired using the Google Street View API. Compared to manual collection, this method offers significant convenience and underscores the potential for large-scale urban deployment. Three pavement engineers, each with over five years of experience, conducted standardized assessments using PSCI criteria; however, variations among their evaluations were also observed. The selected images also capture realistic scenarios, such as different illumination conditions (sunny or overcast) and diverse traffic exposure levels across various road condition ratings.

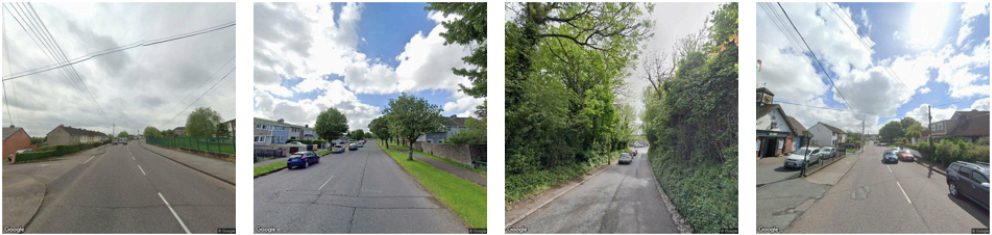

**Figure 3**. Google Street View images from Dataset 2

*2.3 Pavement surface condition index*

Road condition assessment has multiple evaluation criteria, the PSCI was adopted in this study, which is a semi-automated evaluation system that integrates various anomalies quantification with expert visual inspection, as shown in **Table 2**. It categorizes pavement conditions from 1 to 10, with higher ratings indicating better road conditions (**Fig.2**). The PSCI takes various types of surface distresses into consideration, such as cracks, rutting, and potholes.

However, the method faces three critical limitations: (1) resource intensity: extensive training of certified inspectors and corresponding costs is required; (2) logistical constraints: on-site evaluations necessitate road closures, disrupting urban traffic flow and increasing $CO_2$ emissions; (3) hindered scalability: manual data integration delays city-wide condition mapping, hindering time-sensitive maintenance decision-making. Therefore, some studies have adopted deep learning methods to replace manual inspections; however, it also needs extensive data labelling and significant computational resources. Consequently, this motivates our LLM-based method, which employs a zero-shot approach and a deep understanding of the PSCI standard. Through a chain of prompts, this method effectively overcomes the limitations associated with on-site evaluations and the heavy training requirements typical of image-based analyses, demonstrating substantial potential for future research.

**Table 2.** PSCI rating system and treatment measures for asphalt pavement (Mullaney et al. 2014; Qureshi et al. 2023)

| PSCI rating | Primary rating indicators | Secondary rating indicators | Treatment measures | Surface | Structure |
|---|---|---|---|---|---|
| 10 | No visible defects | Road surface in perfect condition | Routine maintenance | Excellent | Very good |
| 9 | Minor surface defects; ravelling or bleeding <10% | Road surface in very good condition | | | |
| 8 | Moderate surface defects; ravelling or bleeding 10% to 30% | Little or no other defects | Resealing and restoration of | Fair | Good |
| 7 | Extensive surface defects; ravelling or bleeding >30% | Little or no other defects; old surface with aged appearance | Skid resistance | Poor | Good |
| 6 | Moderate other pavement defects; other cracking <20%; patching generally in good condition; surface distortion requiring some reduction in speed | Surface defects may be present; no structural distress | Surface restoration | Fair | Fair |
| 5 | Significant other pavement defects; other cracking >20%; patching in fair condition; surface distortion requiring reduction in speed | Surface defects may be present; very localized structural distress (< 5m$^2$ or a few isolated potholes) | Carry out localized repairs and treat with surface treatment or thin overlay | Poor | Fair |
| 4 | Structural distress present; rutting, alligator cracking or poor patching for 5% to 25%; short lengths of edge breakup or cracking; frequent potholes | Other defects may be present | Structural overlay | Poor overall | Poor overall |
| 3 | Significant areas of structural distress; rutting, alligator cracking or poor patching for 25% to 50%; continuous lengths with edge breakup or cracking; more frequent potholes | Other defects may be present | Required to strengthen road; localized patching and repairs are required prior to overlay | Poor overall | Poor overall |
| 2 | Large areas of structural distress; rutting, alligator cracking or very poor patching for >50%; severe rutting (> 75 mm); extensive very poor patching; many potholes | Very difficult to drive | Road reconstruction | Very poor overall | Very poor overall |
| 1 | Extensive structural distress; road disintegration of surface; pavement | Severe deterioration; virtually undriveable | Needs full-depth reconstruction | Failed overall | Failed overall |

| failure; many large and deep potholes; extensive failed patching | | with extensive base repair |

## 2.4 LLM-based model development

### 2.4.1 Large language models

In recent years, large language models (LLMs) based on attention mechanism architectures (Vaswani et al. n.d.) have achieved revolutionary breakthroughs, exemplified by models such as ChatGPT, Gemini, and DeepSeek. With ever-increasing training parameters and vast amounts of internet data, LLMs demonstrate tremendous potential in understanding and generating both short and long texts, which has also accelerated scientific research.

For example, when we upload an image of a road to an LLM and ask, "Is there a crack on the road?", the task becomes multimodal, as illustrated in **Fig.4**. The process involves several stages, beginning with pre-processing, the image is analyzed using a pretrained vision model to extract key features—such as road color and the presence of cracks—which are transformed into a numerical feature vector, while the text is tokenized and converted into a high-dimensional vector through embeddings, with its semantic meaning refined by an encoder. These vectors are then aligned via a cross-modality transformer that projects them into a shared high-dimensional space. During semantic analysis, self-attention mechanisms assess the relevance of various elements, assigning higher weight to critical terms like 'crack' and enabling the model to infer whether a crack exists on the road. Finally, in the decoding stage, the numerical inference is converted back into natural language, resulting in a response such as "Yes, there is a crack on the road" with a confidence level of 90%. Throughout this process, detailed and structured prompts guide the LLM to perform precise semantic analysis, yielding accurate and contextually appropriate responses.

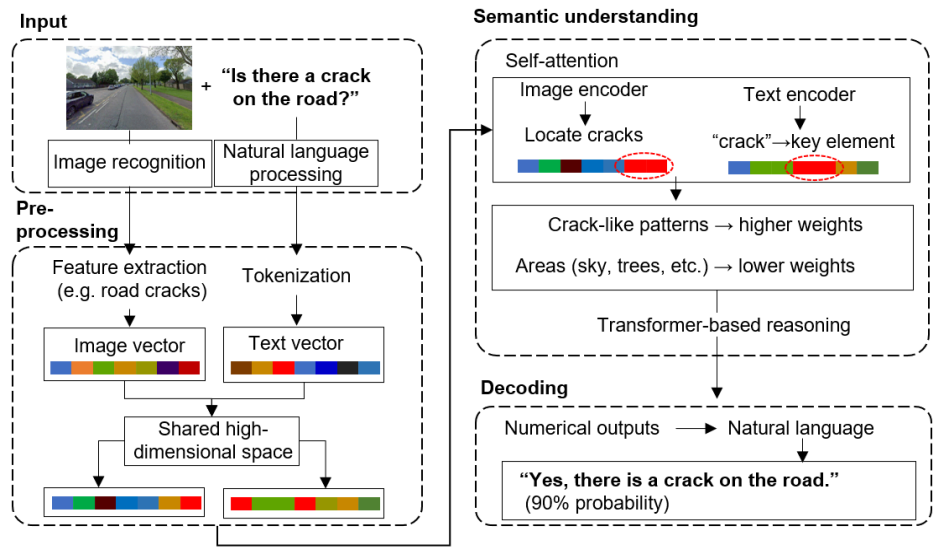

**Figure 4**. LLMs workflow for multimodal task

### 2.4.2 Multi-model development for PSCI rating

Prompt engineering is the process of refining the input by providing precise and accurate descriptions to guide an LLM, ensuring it generates the most relevant, accurate, and high-quality responses for a specific task. It involves wording, structure, and context to optimize the LLM's performance on specific tasks. Prompt engineering has become a pivotal technique for optimizing the performance of LLMs, such as OpenAI's GPT (OpenAI et al. 2024). Unlike traditional supervised machine learning and deep learning methods, which require extensive specific datasets, rigorous training processes, and parameter optimization, LLMs are general models that are able to operate across different domains. This enables them to adapt to new tasks effortlessly in a few or zero-shot learning fashion. As analyzed before, road surface condition assessment ML models traditionally rely on collecting and annotating large datasets, training specialized algorithms, and fine-tuning hyperparameters—processes that are both time-consuming and resource-intensive. By contrast, the LLM-based method proposed in this study leverages zero-shot learning through strategic prompt design, achieving comparable outcomes without the need for extensive data preparation or manual calibration. This makes it a scalable and efficient solution for infrastructure monitoring tasks and hints at its potential to be transferred to different road assessment indexes.

This study developed five progressively sophisticated models (Model 1 to Model 5), each incorporating a unique combination of strategies inspired by OpenAI's prompt design principles (OpenAI et al. 2024). The strategies (S) include adopting a persona to enhance contextual understanding (S1), providing detailed queries for improved task relevance (S2), using delimiters to clearly separate input components (S3), specifying explicit step-by-step instructions for task completion (S4), and designing comprehensive prompts to ensure consistency and robustness (S5). The step-by-step guidance is an approach based on the Chain-of-Thought prompting technique (Chia et al. 2023). Each of these different strategies can be deployed with different levels of complexity, requiring more prompting. **Table 3** shows which strategies are used by each model, with their complexity and depth progressively increasing (more checkmarks indicate a more detailed and extensive description of the corresponding strategy). Model 1 serves as a baseline, utilizing basic strategies such as detailed queries (S2), delimiters (S3), and comprehensive prompts (S5) but omitting contextual adaptation (S1) and step-by-step guidance (S4). Therefore, Model 1 only includes a simple PSCI standard and basic queries. Model 2 introduces S1 to enhance contextual relevance, while Model 3 adds S4 to improve task-specific clarity. In Model 2, the LLM acts as a pavement engineer for judgment, while in Model 3, a more detailed description of the PSCI is provided. Model 4 incorporates all five strategies, offering a comprehensive but less detailed approach compared to Model 5. Finally, Model 5 represents the most advanced application, combining all strategies with maximum depth and precision, resulting in a highly consistent and effective evaluation framework. In Models 4 and 5, we further guide the LLM to assess images following our designed process, with Model 5 including more detailed descriptions in S2 and S4. The process includes the following steps:

(1) The user provides road condition grading criteria.
(2) The user provides an image in Base64 format, focusing solely on the asphalt pavement.
(3) The LLM pays attention to any anomalies on the asphalt pavement, such as surface defects (e.g., ravelling, bleeding), pavement defects (e.g., longitudinal cracks, transverse cracks), and structural distresses (e.g., alligator cracks, rutting, potholes, surface distortion, edge breakup, patching). Anomalies caused by shadows or water stains are excluded.
(4) The LLM estimates the proportion of each type of anomaly on the asphalt surface, providing more specific instructions in Model 5 for improved clarity, which forms the basis for subsequent road classification.
(5) The LLM determines the grade of the road surface using the provided grading criteria, with more detailed queries in Model 5 to enhance precision.
(6) The LLM responds with a single numerical value for the grade, hiding the thought process.

**Table 3.** Strategies used across models

| Strategies | Model 1 | Model 2 | Model 3 | Model4 | Model 5 |
| --- | --- | --- | --- | --- | --- |
| S1 | · | ✓ | ✓ | ✓✓ | ✓✓ |
| S2 | ✓ | ✓ | ✓✓ | ✓✓✓ | ✓✓✓✓ |
| S3 | ✓ | ✓ | ✓ | ✓ | ✓ |
| S4 | · | · | · | ✓✓ | ✓✓✓ |
| S5 | ✓ | ✓ | ✓ | ✓ | ✓ |

Note: Strategies S1-S5 are derived from OpenAI's official guidelines for prompt engineering. The checkmarks (✓) indicate the use of a particular strategy in the respective model, while the cross (✗) indicates the strategy was not used. More checkmarks represent a more extensive description of the strategy.

*2.5 Model benchmarking with multi-level assessors*

To the best of our knowledge, no machine learning regression approach in the literature can assess the PSCI rating of a given picture without an extensive labeled training dataset. Given the lack of such a dataset for PSCI ratings, we opted to compare our approach to human assessors. We provided road condition assessment guidelines and the PSCI levels description to participants with different levels of expertise in road condition evaluation, asking them to assess the two datasets presented above. This diverse evaluation process, involving participants with varying levels of expertise, helped establish a comprehensive benchmark for the model's performance evaluation. Participants were grouped into categories including 'Expert' (3 participants), 'Intermediate' (6 participants), and 'No Experience' (12 participants) based on their experience in road assessment and civil engineering. The 'Expert' category consisted of individuals with deep insights and professional experience with over 5 years in this field, while the 'No Experience' category included participants without any background in the subject area. The 'Intermediate' category represented those with some familiarity (e.g., a background in civil engineering) but no practical experience in road assessment at the industrial level. This diversity provided a comprehensive evaluation, balancing expert insights with diverse perspectives to validate the model's performance. LLM output includes an aleatoric part; this results in models giving slightly different responses to the same prompt. To account for this variability, each model presented herein has been run 10 times. We considered each run as a separate assessor.

## 3 Model selection and application evaluation

*3.1 Comparative analysis of model performance*

*3.1.1 Evaluation metrics*

This section assesses the performance of the proposed approaches against the human evaluators. The ablation study allows us to appreciate the importance and contribution of the different prompting approaches. We compare the models in terms of Mean Absolute Error (MAE), Mean Squared Error (MSE), and Pearson Correlation Coefficient (referred to as "correlation"). In addition, we use Principal Component Analysis (PCA) to gain insight into how different assessors (both human and various models) relate to each other and to the ground-truth ratings. These evaluation metrics and analytical methods provide a comprehensive view of each model's performance, from overall accuracy (MAE, MSE) to alignment in rating trends (correlation), as well as a visualization of the underlying patterns in the data (PCA).

MAE measures the average magnitude of errors between predicted and actual rating, without considering their direction. It is calculated by taking the absolute differences between predictions and actual values and then averaging them. The formula for MAE is shown in equation (1). The $n$ is the number of samples in the dataset, $y_i$ is the rating predicted by the model, and $\hat{y}_i$ is the correct rating. MAE is straightforward and less sensitive to outliers since it does not square the errors, providing a clear sense of the typical error size. MSE measures the average squared differences between predicted and actual rating, as shown in equation 2. Squaring the errors emphasizes larger discrepancies, making MSE more sensitive to predictions that are quite different compared to the correct value, but the units are squared, which can make interpretation less intuitive. The formula for MSE is:

$$MAE = \frac{1}{n}\sum_{i=1}^{n}\left|y_i - \hat{y}_i\right| \qquad (1)$$

$$MSE = \frac{1}{n}\sum_{i=1}^{n}\left(y_i - \hat{y}_i\right)^2 \qquad (2)$$

The Pearson Correlation Coefficient (simply correlation) quantifies the strength and direction of a linear relationship between two variables, with values ranging from −1 to +1. A value of +1 indicates a perfect positive linear correlation, −1 indicates a perfect negative linear correlation, and 0 indicates no linear relationship. Unlike MAE and MSE, Pearson's correlation does not measure accuracy but instead focuses on how closely the predictions follow a linear pattern. The correlation shows if assessors tend to agree on similar marks. PCA has been widely adopted to reduce the dimensionality of signals used for pavement assessment (Ghasemi et al. 2019; Golmohammadi et al. 2025).

*3.1.2 Model selection*

In this experiment, we aim to assess the accuracy of images with an assigned PSCI rating (Dataset 1 presented in section 2.2). **Table 4** presents the MAE, MSE, and correlation for the respective models between the predicted road levels and the ground truth (official PSCI rating) for various models and participant groups. Overall, Models 4 and 5 exhibit the lowest MAE values at 1.03 and 1.07, respectively, demonstrating the highest accuracy in predictions compared to the other models and participant groups. In contrast, Model 1 shows the poorest performance in terms of both MAE and MSE, indicating a significant deviation from the ground truth. Moreover, the intermediate and no-experience participant groups performed relatively poorly on the correlation metric.

Table 4. MSE, MAE, and correlation on Dataset 1

| Models | MAE | MSE | Correlation |
|---|---|---|---|
| Model4 | 1.03 | 1.95 | 0.89 |
| Model5 | 1.07 | 2.12 | 0.85 |
| Expert | 1.10 | 2.39 | 0.86 |
| Model3 | 1.22 | 2.44 | 0.87 |
| Model2 | 1.34 | 2.71 | 0.87 |
| Intermediate | 1.38 | 4.14 | 0.79 |
| No experience | 1.66 | 4.72 | 0.71 |
| Model1 | 2.17 | 6.55 | 0.85 |

**Fig.5** shows the box plot of the MAE and MSE. The *y*-axis represents the MAE and MSE values, where lower values indicate better alignment with the ground truth. The narrow distribution with minimal spread for Model 4 and Model 5 indicates their ability to produce consistent results, reflecting robustness and stability. The expert participants also exhibit relatively low MAE values, indicating their capability to assess road damage accurately. Surprisingly, their performance is slightly inferior to that of the best LLM-based models, even if these have never been trained for this specific task. Conversely, intermediate and inexperienced participants show higher MAE values, suggesting greater deviation from the true labels. Model 1, in particular, has the highest MAE, highlighting the low performance of an LLM model prompted incorrectly. In particular, the superior performance of Model 4 and Model 5 over the other LLM-based models underlines the effectiveness of sophisticated prompt engineering strategies. The 'No Experience' group also shows a wide distribution, indicating significant variability in their assessments, likely due to a higher number of participants and their differing levels of understanding of the PSCI standards. Unsurprisingly, the performance of the human assessors is in line with their experience levels, with only a few people in the 'Intermediate' group outperforming 'Expert' assessors. Even Model 3 outperforms the set of experts based on MSE. The MSE tends to penalize errors more strongly than the MAE. This further highlights the robustness and consistency of the best LLM-based approaches. The plots contain a few limited outliers; the clearest one is an assessor of 'Intermediate' experience that exhibits an MAE of 2.4. We consider this point an outlier and remove it for the following analysis. All models are well-correlated with the correct ratings; in particular, LLMs show the highest correlation values, creating ratings that are more related to the corrected values.

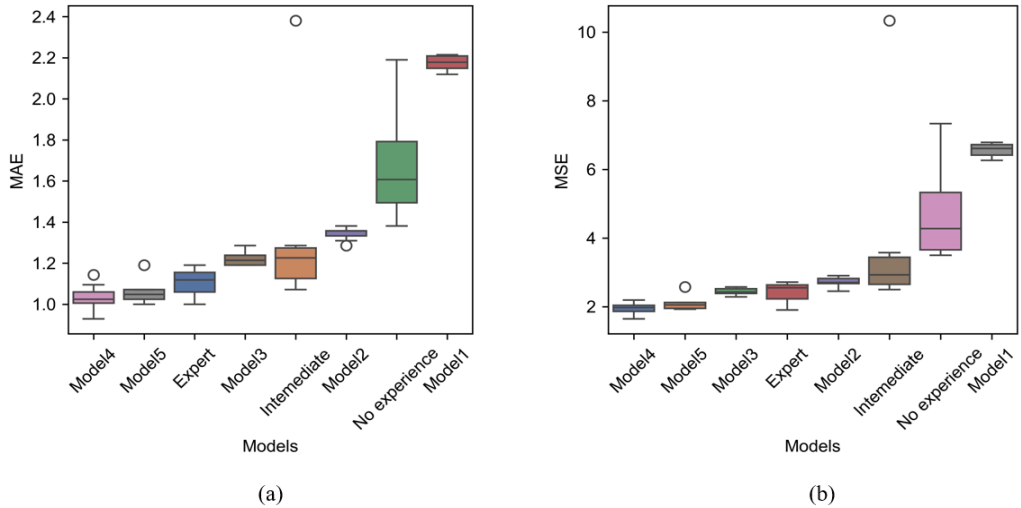

**Figure 5**. (a) MAE and (b) MSE for predictions across different experience levels and models

PCA is a dimensionality reduction technique that identifies the directions (principal components) along which data varies the most. It works by transforming the original variables into a new set of uncorrelated variables ordered by the amount of variance they explain. PCA is commonly used to simplify datasets, reduce noise, and visualize high-dimensional data while retaining as much meaningful information as possible. **Fig.6** presents the results of a PCA used to observe the differences between various assessors. The *x*-axis and *y*-axis represent the first and second principal components (PC1 and PC2), respectively. It is evident that LLM models, particularly Models 4 and 5, are positioned closest to the ground truth along both PC1 and PC2 dimensions, indicating a pattern more similar to the correct labels compared to human participants. This suggests that the LLM models are better aligned with the established standards, especially regarding key features of the road condition assessment. However, while a similar rating pattern to the correct labels is generally indicative of better alignment, it does not always equate to complete correctness. The models' proximity to the ground truth in the PCA space indicates a high level of similarity in overall rating behavior, but certain nuanced aspects may still require human judgment. For example, the evaluations made by participants, particularly the experts, do not deviate significantly along the PC1 dimension, showing that their general understanding of the data is comparable to the LLM models. Yet, the observed variance along PC2 suggests that specific aspects of the road assessment—such as subtle qualitative judgments—may lead to differences between human assessments and LLM predictions.

The closeness of all the points of the different LLM-based models confirms the consistency of the approaches. The progressive improvement brought by the enhanced prompt strategies is clear. Model 1 (the simplest prompt) is clustered far away from the correct values. With the increase in the models' complexity, the clusters of points become closer and closer to the ground truth. The human participants, particularly those with no experience, show wider divergence along PC2, highlighting variability in their assessments. This emphasizes that while advanced LLM models can provide highly consistent evaluations, they are less influenced by subjective differences seen among human participants, which is both an advantage in reducing inconsistency and a limitation in capturing human-expert nuances.

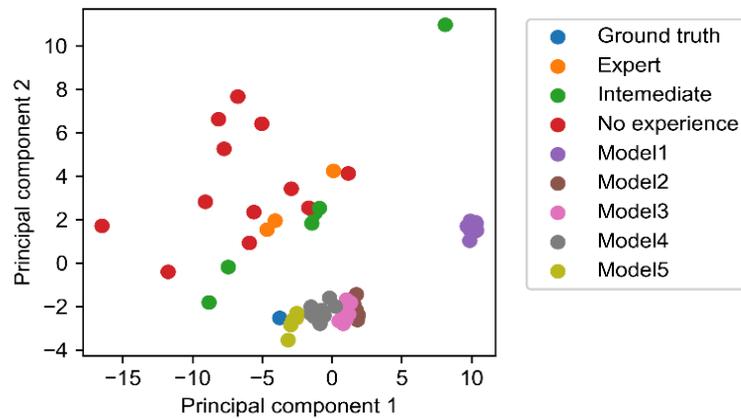

**Figure 6**. PCA analysis of differences across different experience levels and models

*3.1.3 Comparison of selected models*

**Fig.7** illustrates the percentage distribution of road condition ratings (1-10) given by Model 4, Model 5, and expert participants. Model 4 demonstrates a notable concentration around specific ratings, particularly at Rating 6, suggesting a bias or a lack of versatility in assessing a range of road conditions. In contrast, Model 5 displays a more balanced distribution of ratings, closely mirroring the evaluations provided by experts. The expert evaluations, represented in blue, are more evenly spread across all ratings, indicating a nuanced understanding of road conditions. Model 5's distribution aligns more closely with the experts, showing that the sophisticated prompt engineering in Model 5 results in greater reliability and adaptability across various conditions. On the other hand, Model 4's uneven distribution highlights the need for improved prompt strategies, particularly to handle a broader range of road surface conditions more effectively.

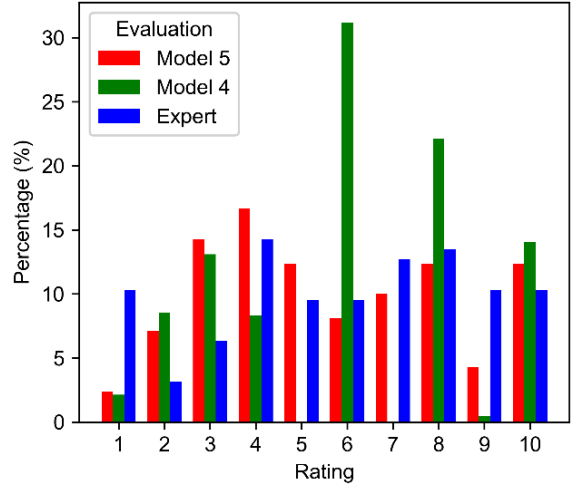

**Figure 7.** Percentage distribution of road condition ratings (1-10) evaluated by Model 4, Model 5, and 'Expert' participants

*3.2 Google Street View images application evaluation*

*3.2.1 ICC-driven expert reference standard*

To further validate the applicability of our model and its potential for large-scale deployment, we extended our analysis to include Google Street View images (see **Table 1,** datasets2). In this study, experienced pavement engineers rated the images according to the PSCI standard to serve as the basis for model evaluation. However, due to differences in engineering experience and focus, some variability was observed among the three experts' ratings. To quantify the consistency among these ratings, we computed the intra-class correlation coefficient (ICC), which measures the proportion of total variance attributable to true differences among subjects relative to measurement error. A higher ICC value indicates that most of the variation comes from actual differences, meaning the ratings are more reliable. **Table 5** shows the ICC results computed for single raters using our dataset. The results indicate that relying solely on one expert may be suboptimal, suggesting that an averaged rating from multiple experts would provide a more robust reference standard.

**Table 5.** ICC results for single raters

| Type | Description | ICC |
| --- | --- | --- |
| ICC1 | Single raters absolute | 0.835558 |
| ICC2 | Single random raters | 0.838619 |
| ICC3 | Single fixed raters | 0.888228 |

Note: ICC1- Assesses exact numerical agreement among raters; ICC2-Assumes raters are randomly drawn from a larger population; ICC3-Evaluates consistency among a specific set of raters.

Subsequently, we calculated the ICC for all possible two-expert combinations and for the three-expert combination using a fixed raters model—where ICC3 represents the reliability of individual (single) ratings, and ICC3k represents the reliability when using the mean rating. The results are summarized in **Table 6**. Notably, the combination of Expert-2 and Expert-3 yielded the highest reliability (ICC3 = 0.928889 and ICC3k = 0.963134). Although the three-expert combination produced an ICC3 of 0.888228 and an ICC3k of 0.959743, based on the highest reliability indices, we ultimately selected the average rating of Expert-2 and Expert-3 as the reference standard for evaluating the application of our model to Google Street View images.

Table 6. ICC Results for Two‑ and Three‑Expert Combinations

| Combination | Raters | ICC3 (Single) | ICC3k (Mean) |
| --- | --- | --- | --- |
| Expert-1, Expert-2 | 2 | 0.863430 | 0.926710 |
| Expert-1, Expert-3 | 2 | 0.869306 | 0.930084 |
| Expert-2, Expert-3 | 2 | 0.928889 | 0.963134 |
| Expert-1, Expert-2, Expert-3 | 3 | 0.888228 | 0.959743 |

*3.2.2 Performance evaluation on Google Street View*

The evaluations are benchmarked against expert assessments (Expert-2 and Expert-3), which serve as the reference standard. **Fig.8a** presents the MAE values for the different models and participant groups. Lower MAE values indicate closer alignment with the expert benchmark. Models 3, 4, and 5 demonstrate the lowest MAE, suggesting that they are the most accurate in predicting road conditions compared to expert evaluations. The narrow range of their box plots further indicates low variability, which implies consistent performance across the set of images. The participant groups, particularly those with no experience, show significantly higher MAE values. The 'No Experience' group exhibits substantial variability, as indicated by the wide spread of the box plot, reflecting inconsistency in their evaluations. Intermediate participants also show higher error values compared to the models, though with slightly less variability than the inexperienced group. This variability highlights the challenges associated with human subjectivity, especially when expertise is lacking. **Fig.8b** displays the MSE values for the same models and participant groups. Similar to the MAE results, Models 3, 4, and 5 achieve the lowest MSE values, reinforcing their superior performance in comparison to other models and participant groups. The narrow variability in their box plots highlights their robustness in maintaining low error rates across various assessments.

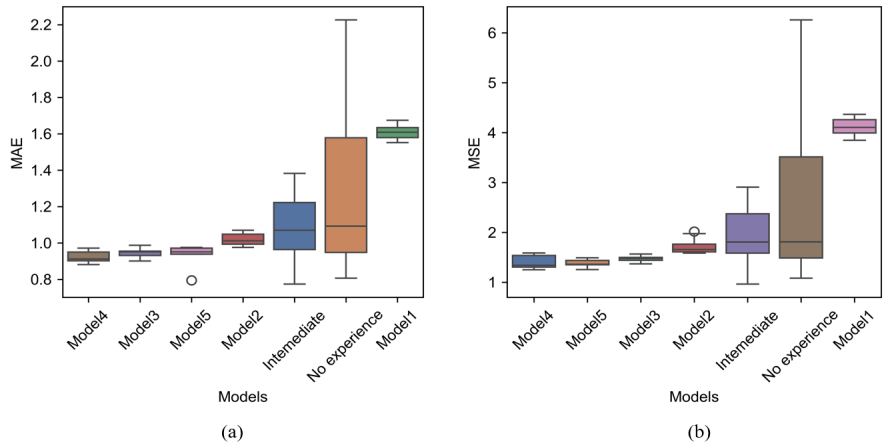

**Figure 8.** (a) mean Absolute Error (MAE) and (b) mean Squared Error (MSE)

**Fig.9a** provides a visualization of the PCA results based on GSV images. Overall, the predictions from the models and the human evaluations both exhibit concentrated patterns, forming two clearly clusters. The two separate clusters indicates notable differences between model-based and human assessments, particularly regarding the interpretation of certain detailed road conditions. Models 4 and 5 closely align with expert evaluations along the PC1 axis. Intermediate participants are scattered but are still closer to the experts compared to those without experience, suggesting some level of alignment but lacking the precision achieved by models with more sophisticated prompts. The 'No Experience' group shows considerable distance from the expert benchmark, reflecting their inconsistent and varied assessments.

**Fig.9b** depicts the percentage distribution of road condition ratings (1-10) provided by Model 3, Model 4, Model 5, and expert participants. Model 3 and 4 demonstrate a concentration of evaluations at specific rating levels, particularly Rating 6, which suggests a bias or limitation in prompt handling across a full range of conditions. Model 5, by contrast, exhibits a more balanced distribution, which closely resembles the expert evaluations and reflects the robustness of its prompt engineering in consistently evaluating a variety of conditions. Experts, represented in blue, show a relatively even distribution across all rating levels, highlighting their nuanced understanding of road conditions. The rating distribution highlights the importance of balanced evaluations. Model 5's close resemblance to expert evaluations underscores the effectiveness of its prompt engineering, which enables it to handle diverse conditions without significant bias.

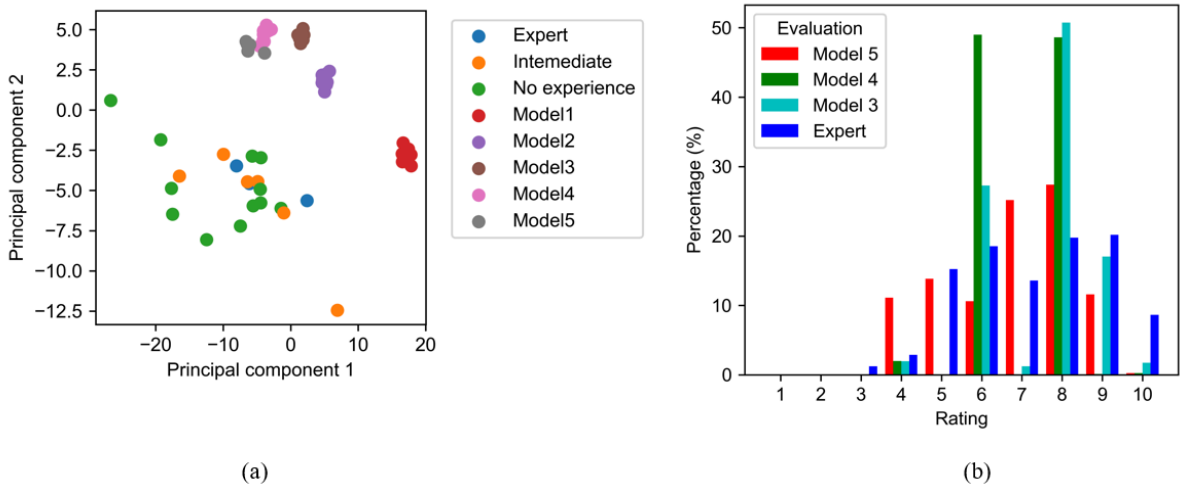

**Figure 9.** (a) PCA and (b) rating distribution of model and participant

Model 5 was selected as the model that best resembled experts' evaluation. **Fig.10** illustrates the comparison between Model 5 and expert ratings for both Dataset 1 and Dataset 2. The difference between the ratings given by Model 5 and the expert mean for each rating level (1-10) is presented. The box plots highlight the variability in the differences across different rating levels. At lower rating levels (1-3), Model 5 tends to slightly overestimate compared to expert evaluations, with differences predominantly positive. This suggests that Model 5 might be interpreting early signs of degradation more conservatively than the experts. As the rating level increases, the difference reduces and becomes more balanced, indicating better alignment with the expert assessments at higher levels of road quality. The variability in differences is relatively larger at mid-range levels (4-8), which suggests that Model 5 struggles more where road quality is neither extremely poor nor excellent. This variability highlights areas where prompt engineering could be further refined to better handle the subtleties of mid-range conditions. At higher rating levels (9-10), Model 5 systematically underestimates well-maintained road conditions by roughly one rating point, with a median difference of about -1.

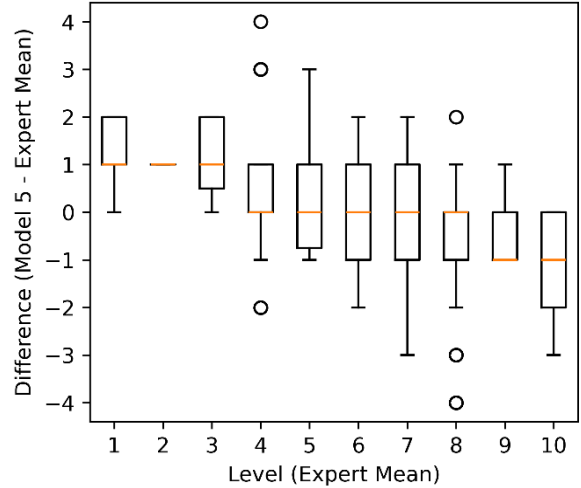

Figure 10. The differences across different rating levels between Model5 and expert

## 4 Discussion

*4.1 Comparison with traditional methods*

**Table 7** summarizes the key differences between traditional road condition assessment methods, supervised machine learning, and the proposed LLM-based approach (Models 4 and 5). Traditional methods typically rely on manual visual inspections, which can lead to inconsistent results due to human subjectivity. Moreover, these methods struggle to adapt to new or varied road conditions because they are constrained by the fixed features they recognise. Supervised machine learning approaches, such as deep learning models for road condition classification and YOLO-based anomaly detection, rely on predefined features, reducing the need for manual feature engineering. However, they require large amounts of labelled data and often struggle to generalize when the training set does not cover all possible conditions. Additionally, these models are sensitive to environmental factors like lighting and weather, function as "black boxes" with limited interpretability, and YOLO-based systems can miss small anomalies or fail to capture irregular shapes accurately. In contrast, LLM-based models, particularly Models 4 and 5 in this study, leverage advanced prompt engineering and zero-shot learning to achieve enhanced adaptability and accuracy in road condition assessment. These models are capable of processing diverse inputs and responding to context-specific prompts, allowing them to generalize across different conditions. The adaptability of LLMs makes them highly suitable for the complex and dynamic nature of road condition evaluation, where the variations in road surfaces, environmental factors, and damage types demand a flexible approach. Overall, the LLM-based approach, enhanced by prompt engineering, offers a robust and consistent solution.

**Table 7**. Key differences between traditional methods, supervised machine learning and the LLM-based approach

| Dimension | Traditional methods | Supervised machine learning | LLM-based approach |
| --- | --- | --- | --- |
| Evaluation method | Manual inspection | Fixed-feature automated analysis requiring large labeled datasets | Advanced prompt engineering with context-specific processing |
| Consistency | Subject to human subjectivity; variable results | Consistently aligns with expert evaluations | Consistently aligns with expert evaluations |
| Adaptability | Highly adaptable to diverse and evolving road conditions and standards | Limited to predefined features; not responsive to standards and conditions changes | Highly adaptable to diverse and evolving road conditions and standards |
| Feature dependency | Relies heavily on manual inspection and extensive training of personnel | Requires dedicated feature engineering | Uses pre-learned features that are part of the LLM model |
| Scalability | Labor-intensive | Efficiently scalable | Efficiently scalable |

*4.2 Limitations and future directions*

Prompt engineering is essential for optimizing LLMs in road condition assessment in this study. By incorporating explicit, clear instructions and step-by-step guidance, Models 4 and 5 significantly outperform simpler models (like Model 1), achieving accuracy and consistency that often match or exceed human evaluations. Despite these improvements, errors are most common in mid-range ratings—borderline cases where subtle differences in road conditions make classification difficult for both LLMs and human experts. This indicates that further refinements are needed.

**Fig.11** presents several prompt engineering techniques: zero-shot, in-context learning (ICL), chain of thought (CoT), and tree of thought (ToT). Zero-shot queries an LLM without examples, while ICL (few-shot learning) provides sample Q&A pairs (Brown et al. 2020). CoT incorporates step-by-step reasoning for complex problems (Wei et al. 2023), further enhanced by self-consistency through multiple reasoning paths and majority voting (X. Wang et al. 2023). When one reasoning path falls short, ToT uses multiple decision paths to explore diverse outcomes. In this study, we introduced CoT-style prompts (Models 4 and 5) to enhance the accuracy and consistency of road condition assessments. However, due to current API limitations, we could not provide examples of labeled road images directly in the prompt. Future research can address this gap by incorporating high-quality images and refined evaluation criteria to offer richer, more concrete guidance. Additionally, exploring more advanced techniques—such as multi-path ToT or self-consistency—could further improve the model's performance, particularly in borderline classification cases.

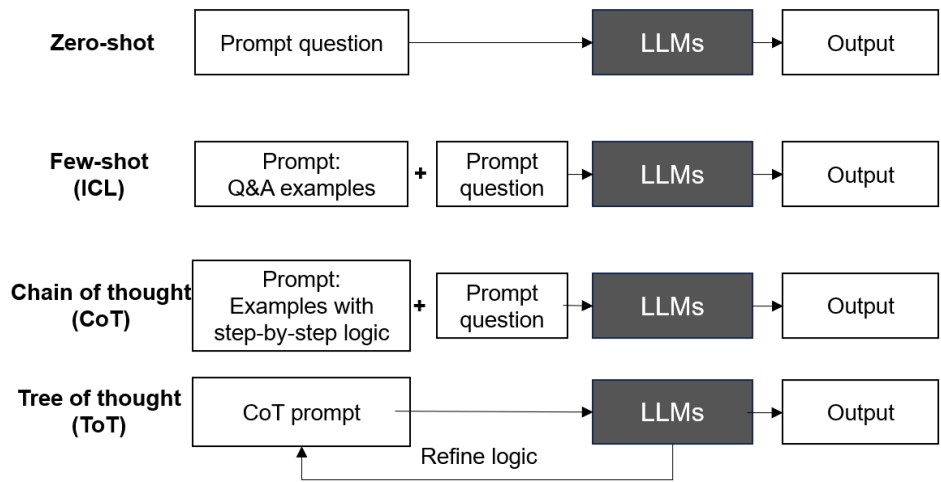

**Figure 11.** Types of prompt engineering techniques(Wu et al. 2024)

# 5 Conclusion

This study innovatively leverages large language models (LLMs) integrated with advanced prompt engineering strategies for automating pavement surface condition assessments. The results indicate that comprehensive prompt engineering, as implemented in Model 5, significantly enhances the model's ability to provide accurate and consistent evaluations. These evaluations closely align with expert assessments and surpass human performance.

Key advancements include the ability of LLMs to adapt to diverse road conditions using zero-shot learning frameworks, eliminating the need for extensive labeled datasets. By leveraging techniques such as step-by-step instructions, contextual prompts, and detailed task specifications, the proposed models demonstrated robust performance across varying scenarios. Principal Component Analysis (PCA) and error metrics such as MAE (1.07) further validated the effectiveness of the approach, particularly in reducing variability and improving consistency compared to human evaluators (MAE 1.10). Moreover, the optimized model showed strong agreement with expert assessments (MAE 0.93) when evaluating Google Street View road images, underscoring its potential for large-scale future applications.

Challenges remain, particularly in mid-range road condition ratings (e.g., PSCI 4–7), where nuanced features require more precise contextual understanding and differentiation. To address these challenges, future work should integrate high-quality road imagery, refined evaluation criteria, and advanced prompt engineering techniques—such as multi-path reasoning and self-consistency—to better capture and differentiate subtle road features. This research lays the foundation for scalable, efficient, and reliable infrastructure monitoring systems. By addressing the limitations and leveraging the adaptability of LLMs, future studies can pave the way for widespread adoption in urban planning and maintenance.

# Acknowledgement

This publication has emanated from research conducted with the financial support of the EU Commission Recovery and Resilience Facility under the Science Foundation Ireland Future Digital Challenge Grant Number 22/NCF/FD/10932, Science Foundation Ireland Frontiers for the Future Programme, 21/FFP-P/10090, and by SFI, Grant number 12/RC/2289-P2 co-funded under the European Regional Development Fund. For the purpose of Open Access, the